\pdfoutput=1

\documentclass[11pt]{article}

\usepackage{EMNLP2022}

\usepackage{times}
\usepackage{latexsym}

\usepackage[T1]{fontenc}

\usepackage[utf8]{inputenc}

\usepackage{microtype}

\usepackage{inconsolata}

\usepackage{graphics}
\usepackage{graphicx}
\usepackage{amsmath}
\usepackage{amsfonts}
\usepackage{mathbbol}
\usepackage{multirow}
\usepackage{array}
\usepackage{booktabs}
\usepackage{algpseudocode}
\usepackage{algorithm}

\title{Syntactically Robust Training on Partially-Observed Data for Open Information Extraction}

\author{Ji Qi$^1$, Yuxiang Chen$^2$, Lei Hou$^1$, Juanzi Li$^1$, Bin Xu$^{1}$\thanks{\, Corresponding author: xubin@tsinghua.edu.cn} \\
  $^1$Department of Computer Science and Technology, BNRist, \\
      Tsinghua University, Beijing, 100084, China \\
  $^2$University of California, San Diego \\
  \texttt{qj20@mails.tsinghua.edu.cn}, \texttt{yuc129@ucsd.edu}
}

\begin{document}
\maketitle
\begin{abstract}
Open Information Extraction models have shown promising results with sufficient supervision.
However, these models face a fundamental challenge that the syntactic distribution of training data is partially observable in comparison to the real world.
In this paper, we propose a syntactically robust training framework that enables models to be trained on a syntactic-abundant distribution based on diverse paraphrase generation.
To tackle the intrinsic problem of knowledge deformation of paraphrasing, two algorithms based on semantic similarity matching and syntactic tree walking are used to restore the expressionally transformed knowledge.
The training framework can be generally applied to other syntactic partial observable domains.
Based on the proposed framework, we build a new evaluation set called CaRB-AutoPara, a syntactically diverse dataset consistent with the real-world setting for validating the robustness of the models.
Experiments including a thorough analysis show that the performance of the model degrades with the increase of the difference in syntactic distribution, while our framework gives a robust boundary.
The source code is publicly available at \href{https://github.com/qijimrc/RobustOIE}{https://github.com/qijimrc/RobustOIE}.
\end{abstract}

\section{Introduction}

Open Information Extraction (OpenIE) involves converting natural text to a set of n-ary structured tuples of the form (arg$_1$, predicate, arg$_2$, ..., arg$_n$), composed of a single predicate as well $n$ arguments.
With the advantages of domain independence and scalability, OpenIE serves as a backbone in natural language understanding and fosters many applications such as text summarization~\cite{fan-etal-2019-using} and question answering~\cite{yan2018assertion}.

\begin{figure}[h]
    \centering
    \includegraphics[scale=0.5]{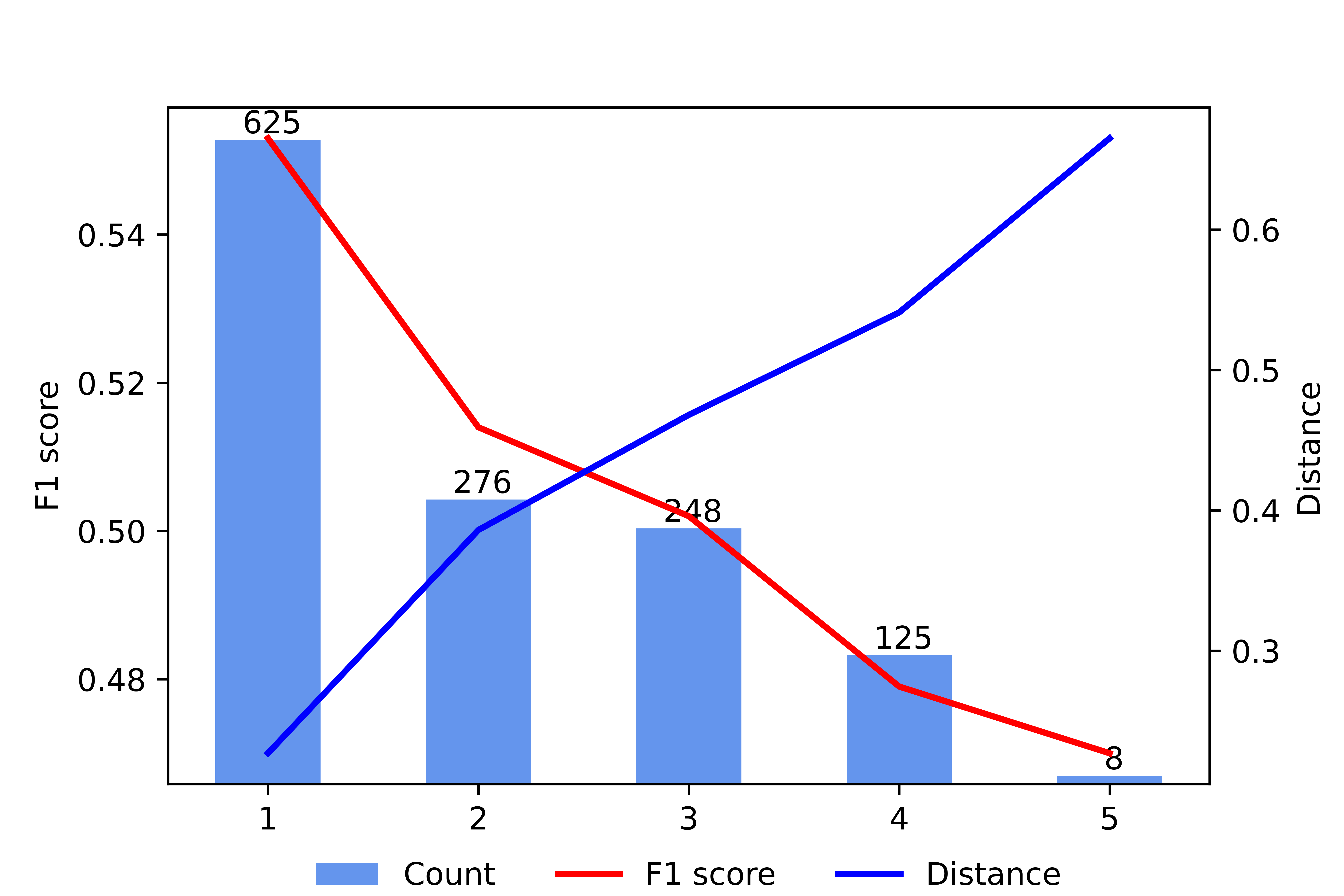}
    \caption{Cluster CaRB into 5 subsets based on the HW-Syntactic Distance and evaluate the IMOJIE model on them. The horizontal axis indicates the indices sorted by the number of samples (above the bars) in the subsets. The left and right vertical axes represent the F1 scores of the model and the distance between the training set and the clustering center of each subset, respectively.}
    \label{fig:example}
\end{figure}

Tremendous efforts have been devoted to build models that can better fit the extractions from texts~\cite{banko2007open,angeli-etal-2015-leveraging,saha2018open,kolluru-etal-2020-imojie,yu2021maximal}.
However, a major issue remaining in OpenIE is the syntactic partial observability -- the syntactic distribution on the existing training set is only based on partial observations, and it is far from covering the entire syntactic hypothesis space in the real world.
This issue creates a challenge that the models rely heavily on the syntactic forms during training, and degrade significantly when the syntactic distribution changes in the real world.

An evaluation is shown in Figure~\ref{fig:example}. We cluster the CaRB~\cite{bhardwaj2019carb} samples based on the \textit{HW-Syntactic Distance} (introduced in Sec.~\ref{subsec:analysis}), which is an effective metric that measures the syntactic difference between two sentences, and evaluate the state-of-the-art model trained on the OpenIE4 data~\cite{kolluru-etal-2020-imojie} on them.
A frustrating result shows that the model performance exhibits a significant degradation as the syntactic similarity between the training set and clustering centers of subsets decreases.
The biased performance comes from the inconsistency of the syntactic distributions among data. For example, in Figure~\ref{fig:example}, the model achieves a depressing F1 score of 0.47 on the subset 5 with the lowest average syntactic similarity to the training set.
Therefore, to build robust OpenIE systems, we need to train the models on a sufficient syntactic distribution.

However, it is not trivial to obtain data that are both diverse and accurate to satisfy the distribution assumption.
First, it is extremely expensive and almost impossible for human annotators to provide a large corpus with diverse syntactic expressions. Second, existing distant supervision-based methods are not applicable to OpenIE due to the uncertainties of both the type and form of arguments and predicates.

Humans learn syntactic grammar by paraphrasing the same meaning into different expressions. For example, the following two sentences convey the same meaning in different syntactic forms. The diverse paraphrases of normal-scale training data can guarantee sufficient syntactic distribution.
However, an intrinsic problem that hinders the efficiency of this approach is the \textbf{Knowledge Deformation}. In the following example, it is difficult to reveal the source object \textit{Earth} in the target paraphrase $b$ as it has been transformed into the form of \textit{the name of the planet} with different syntax.

\begin{itemize}
\item \textit{a. After five years of searching, the Colonials found a new world and named it Earth.}
\item \textit{b. The colonials searched for five years until they discovered a new world and gave him the name of the planet.}
\end{itemize}

In this paper, we propose a syntactically robust training framework that enables OpenIE models to be trained on a syntactic-abundant distribution based on the diverse paraphrase generation.
Specifically, we first generate a large-scale syntactically diverse paraphrase candidates set for the training data based on an off-the-shelf paraphrase generator.
Then, we propose two adaptive algorithms to recover the deformed arguments of the original knowledge, a semantic similarity-based matching method to locate the disordered arguments and a syntactic tree walking-based method to complete the consecutive spans.
We further employ the generative T5~\cite{raffel2020exploring} model to restore the deformed predicates as there are potential tense and voice changes in the target paraphrase.
Finally, a simple but effective denoising method is utilized to prevent the impact of false positives in training.

To exhaustively validate the syntactic robustness of OpenIE models in the real-world setting, an additional evaluation set including diverse paraphrases and knowledge triples has been built on the basis of CaRB.
We conduct experiments on the standard and our proposed evaluation sets based on the division of different syntactic categories, and a comprehensive analysis shows that the model performance decreases with increasing the difference in the syntactic distributions, while our training framework gives a robust boundary.

\section{Syntactically Robust Training Framework for OpenIE}

\subsection{Overview}

The task of OpenIE aims to build a model $\textrm{p}_{\theta}$ to automatically extract a set of n-ary tuples $\{r_i=(a_1, p_r, a_2, a_3, ..., a_n)\}_{i=1}^m$ for each sentence, where $p_r$ indicates the predicate, $a_1, a_2$ indicate the subject and object, and $a_3, ..., a_n$ refer to the other argments such as time and location.
Given a training set $\mathcal{D}=(s_1,s_2, ..., s_{|\mathcal{D}|})$ consisting of sentences samples, where each sentence exhibits a syntactic structure $e^s$.
Our goal is to maximize the expectation of log-likelihood function $\log \textrm{p}_{\theta}(r_1, ..., r_m, e^s | s)$ with respect to the data distribution $\textrm{p}_{\mathcal{D}}$ as following:

\begin{align}
\begin{split}
  \mathcal{L}(\theta) =&\mathbb{E}_{r_i,e^s\sim \textrm{p}_\mathcal{D}}[\log \textrm{p}_{\theta}(r_1, ..., r_m, e^s|s)] \\
                      =&\mathbb{E}_{r_i,e^s\sim \textrm{p}_\mathcal{D}}[\log \textrm{p}_{\theta}(r_1, ..., r_m|e^s, s) \textrm{p}_{\theta}(e^s | s)]
\nonumber
\end{split}
\end{align}

where different OpenIE models may adopt a distinct strategy to model the probability $\textrm{p}_\theta$, such as the triples generating paradigm~\cite{kolluru-etal-2020-openie6} or sequence labeling paradigm~\cite{zhan2020span}, and the maximization process is performed by gradient ascent.

\begin{figure*}[h]
  \centering
  \includegraphics[scale=0.24]{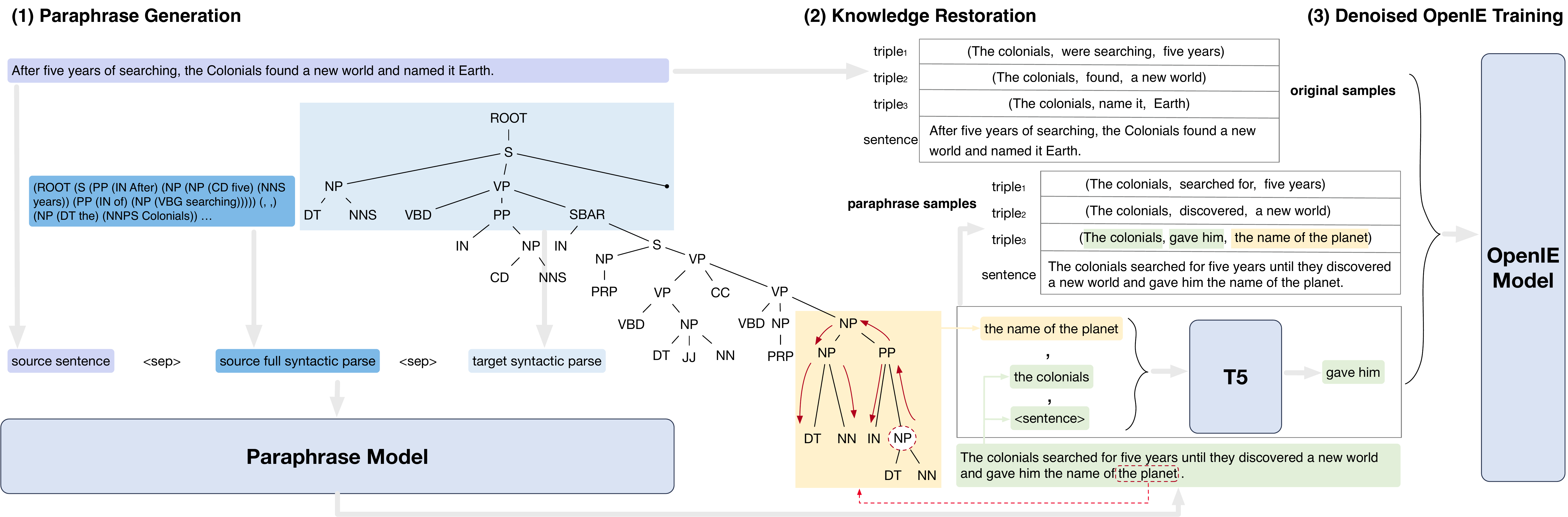}
  \caption{Overview of the proposed framework. Based on the diverse paraphrase candidates set generated by a syntactically controllable model, two algorithms, semantic similarity-based arguments localization and syntactic tree walking, are used to restore the deformed arguments. By taking the arguments as conditions, the predicates are generated with the T5 model.}
  \label{fig:framework}
\end{figure*}

The syntactic distribution in training set $e^s\sim p_\mathcal{D}$ is far from covering the entire syntactic hypothesis space, and plays a fatal role in OpenIE modeling. In this research, we aim to expand the training with a sufficient syntactic distribution. The proposed framework is illustrated in Figure~\ref{fig:framework}. We first generate a syntactically diverse paraphrase candidate set for the training data with an off-the-shelf paraphrase generation model. Then, we restore the deformed arguments using semantic similarity-based matching and syntactic tree walking algorithms, followed by a T5-based predicate restoration. Finally, a denoised training is adopted to optimize the model on the sufficient distribution.

\subsection{Paraphrase Generation}
\label{subsec:paraphrase-generation}

To create syntactically diverse paraphrases candidates set on $\mathcal{D}$, we adopt AESOP~\cite{sun2021aesop}, a syntactically controllable paraphrase generation model as our generator. As can be seen in Figure~\ref{fig:framework}, by utilizing the BART~\cite{lewis-etal-2020-bart} model as a backbone, the model takes \textit{source sentence<sep>source full syntactic parse<sep>target syntactic parse} as the input sequence, and outputs a sequence of the form \textit{target syntactic parse<sep>paraphrase} in which the generated paraphrase conforms with the pruned target syntax.

The AESOP model used in our work is trained on a parallel annotated data with a two-level target syntactic tree.
During generation given the training set $\mathcal{D}$, we first get their constituency parse trees\footnote{We use Stanford CoreNLP~\cite{manning2014stanford}.} $\{T^\mathcal{D}_{s_1}, ..., T^\mathcal{D}_{s_{|\mathcal{D}|}}\}$ and linearize them into parentheses trees as the source full syntactic parses (A part is shown in Figure~\ref{fig:framework}). Then, we collect a set of constituency parse pairs pruned at height 3 $\{(T^\mathcal{P}_{s_1},T^\mathcal{P}_{t_1}), ...,(T^\mathcal{P}_{s_{|\mathcal{P}|}},T^\mathcal{P}_{t_{|\mathcal{P}|}})\}$ from the ParaNMT-50M~\cite{wieting-gimpel-2018-paranmt} and count their frequencies. For each sentence in $\mathcal{D}$, following the original work we obtain $m$ most similar parses $\{T'^{\mathcal{P}}_{s_1}, ..., T'^{\mathcal{P}}_{s_m}\}$ by calculating weighted ROUGE scores between parse strings, and select $k$ top-ranked parses from $\{T^\mathcal{P}_{t_1}, ..., T^\mathcal{P}_{t|\mathcal{P}|}\}$ for each $T'^{\mathcal{P}}_{s_i}$ by a sampling with the distribution of:

\begin{align}
T^\mathcal{P}_t\sim p(T^\mathcal{P}_t | T'^{\mathcal{P}}_{s_i}) = \frac{\#(T'^{\mathcal{P}}_{s_i}, T^{\mathcal{P}}_t)}{\sum_j\#(T'^{\mathcal{P}}_{s_i}, T^\mathcal{P}_{t_j})}
\end{align}

where $\#(T'^{\mathcal{P}}_{s_i}, T^\mathcal{P}_t)$ refers to the count of occurrence in the statistic data. In the end, we generate $k$ paraphrases for each sentence in $\mathcal{D}$. For a tradeoff of quality and quantity, we set $k$ and $m$ to 5 and 2, respectively. As a result, we get the paraphrases candidates set $\mathcal{P}$, which is roughly five times the size of sentences in training set $\mathcal{D}$.

\subsection{Knowledge Restoration}
\label{subsec:knowledge-restoration}

As the paraphrases change the expression form of the original sentence, we need to recover the knowledge of transformed triples.
The difficulty of knowledge restoration lies in two aspects: first, the OpenIE arguments are generally formed as a large span of words, which can be rearranged and rephrased in the target sentence. Second, the syntactic changes also lead to a transformation of tense or voice of verbs in the predicates.
For example in Figure~\ref{fig:framework}, the argument \textit{the Earth} changes its expression and length to become \textit{the name of the planet}, and the predicate \textit{were searching} changes its tense to become \textit{searched for}.

Therefore, we first locate the arguments with the contextualized semantic matching and complete it with syntactic tree walking. Then for each pair of recovered arguments, we restore the corresponding predicate with the T5 model~\cite{raffel2020exploring}.

\subsubsection{Argument Restoration}

As the expressional transformations, it is difficult to get the corresponding arguments in the target paraphrase sentence based on methods like pattern matching. Therefore, we utilize the semantic similarity with BERT~\cite{devlin-etal-2019-bert} to locate the arguments. We first compute the embeddings $\mathbf{h}^s\in\mathbb{R}^{|s|\times d}$ and $\mathbf{h}^t\in\mathbb{R}^{|t|\times d}$ for the source sentence $s$ and target paraphrase sentence $t$, respectively. Then, for a triple $(a^s_1, p_r^s, a^s_2)$ where $a^s_i\rightarrow(l^s_i,r^s_i), p_r^s\rightarrow(l^s_p,r^s_p)$ \footnote{For convenient, we use $l^s_i$ and $r^s_i$ to denote the indices of start word and end word of argument $a_i$  in the sentence $s$.} in the source sentence, we calculate the semantic similarity scores $\mathbf{c}^{a_i},\mathbf{c}^{r}\in\mathbb{R}^{|t|}$ by summing the cosine similarities between each word in $a_i^s,p_r^s$ and target words of $t$:

\begin{align}
\hspace{-10pt}\mathbf{c}^{a_i} = \sum_{j=l^s_i}^{r^s_i}\cos(\mathbf{h}^s_j, \mathbf{h}^t), \mathbf{c}^{r} = \sum_{j=l^s_p}^{r^s_p}\cos(\mathbf{h}^s_j, \mathbf{h}^t)
\end{align}

Next, we merge the consecutive indices of target words whose semantic similarity scores are greater than a threshold $\tau$ to get the resulting candidate spans $\{(l^t_{i1}, r^t_{i1}), ..., (l^t_{im}, r^t_{im})\}$ and $\{(l^t_{p1}, r^t_{p1}), ..., (l^t_{pm}, r^t_{pm})\}$ for $a^s_i$ and $p_r^s$, and the final triplets are obtained by selecting a set of spans with the highest total score and no overlap. By applying this algorithm on $\mathcal{P}$, we get dataset $\mathcal{D}^{\mathcal{P}}$. We refer to the set expanded with this newly built set as $\mathcal{D}^{\Phi}=\mathcal{D}\cup\mathcal{D}^{\mathcal{P}}$.

Though the resulting spans based on semantic similarity matching are accurate in position, we find it incomplete due to the fact that words such as prepositions or adverbs can not be matched effectively by the contextualized embedding. On the other hand, a subtree with NP, QP or NX as the root in the constituency parses represents a continuous phrase fragment. Therefore, we propose to use the syntactic tree walking to further complete the target arguments. Specifically, for each word in span $(l^t_{ij}, r^t_{ij})$, we perform a post-order traversal for the target syntactic tree to effectively find the subtree with NP, QP or NX as the root and containing the the word as a node. We obtain the refined span $(l^{t'}_{ij}, r^{t'}_{ij})$ by replacing the original span (if it covers the original span, otherwise the original span is retained) with the corresponding words of the subtree. Finally, we select the optimal target spans $\{(l^{t*}_{1}, r^{t*}_{1}), ..., (l^{t*}_{n}, r^{t*}_{n})\}$ of all arguments from the refined spans set of each argument by a simple optimality criterion that maintains $n$ spans with the highest similarity without overlaps. We retain the argument restoration as Algorithm \ref{alg:argument-restoration} in detailed.

\begin{algorithm}
\renewcommand{\algorithmicrequire}{\textbf{Input:}}
\renewcommand{\algorithmicensure}{\textbf{Output:}}
\caption{Arguments Restoration}\label{alg:cap}
\begin{algorithmic}[1]
\Require Source/target sentence embeddings $\mathbf{h}^s/\mathbf{h}^t$, source tuple $(a^s_1,p_r^s, ..., a^s_n), a^s_i\rightarrow(l^s_i,r^s_i)$
\Ensure target n-tuple $(a^t_1, a^t_2, ..., a^t_n)$
\State get target constituency parse tree $T^t$
\State subtree roots $\mathcal{T}=\{NP,QP,NX\}$
\State threshold $\tau = 0.7$
\For{each argument $a^s_i \in (a^s_1, ..., a^s_n)$}
\State calculate scores $\mathbf{c}^{a_i} = \sum_{j=l^s_i}^{r^s_i}\cos(\mathbf{h}^s_j, \mathbf{h}^t)$
\State get candidate spans $csp_i=\{(l^t_{i1}, r^t_{i1}), ...\}$ by merging the consecutive indices with values greater than $\tau$ in $\mathbf{c}^{a_i}$
\For {$sp_{ij}=(l^t_{ij}, r^t_{ij}) \in csp_i$}
\For {$tok_k \in sp_{ij}$}
\State $traverse(T^t)$ to find subtree $T^t_{k}$ that satisfies: $T^t_{k}.root\in \mathcal{T}$ \&\& $tok_j\in T^t_{k}$
\State  $T^t_{j} \gets T^t_{j} + T^t_{k}$
\EndFor
\State $sp'_{ij} = (l^{t'}_{ij}, r^{t'}_{ij}) \Leftarrow T^t_j$
\EndFor
\State  $csp'_i=\{sp'_{i1}, sp'_{i2}, ...\}$
\EndFor
\State return $\{sp^*_i | sp^*_i\in csp'_i, i=1,..., n\}$ with highest score without overlaps
\end{algorithmic}
\label{alg:argument-restoration}
\end{algorithm}

\subsubsection{Predicate Restoration}
\label{subsubsec:predicate-restore}

As the paraphrase may change the voice and tense of the predicate in the original sentence, it is not applicable to recover the predicate using the same algorithm as the arguments restoration. We adopt the T5 model~\cite{raffel2020exploring} to restore the predicate in the target paraphrase sentence, as there are a lot of predicates that can not be found from the continuous span of the original sentence. Specifically, we build a new dataset on $\mathcal{D}$ with the same corpus size. For each data sample in the new dataset, the input is of the form of \textit{source sentence, argument$_1$, argument$_2$ <$\backslash$s>}, and the output is a generated sequence referring to the predicate. We train the basic T5 model on the new dataset. Then, we restore the predicate for each pair of arguments obtained from the algorithm~\ref{alg:argument-restoration} to get a final refined set $\mathcal{D}^{\mathcal{P}'}$. We refer to the refined final expanded set as $\mathcal{D}^{\Psi}=\mathcal{D}\cup\mathcal{D}^{\mathcal{P}'}$.

\subsection{Denoised Training}
\label{subsec:denoised-training}

During the training, we aim to maximize the expectation of log-likelihood function with respect to the data distribution:

\begin{align}
\hspace{-10pt} \mathcal{L}(\theta) =&\mathbb{E}_{(r_1, ..., r_m) \sim p_d}[\log p_{\theta}(r_1, ..., r_m|s)]
\end{align}

where $p_d$ refers to a training set, and $p_\theta$ is a neural network model with learnable parameters $\theta$, which either employs the sequence labeling paradigm to predict classification labels on the input sequence, or leverages the generative paradigm to generate target triples each token at a time. In this paper, we validate our proposed training framework on IMOJIE~\cite{kolluru-etal-2020-imojie}, a strong generative model that predicts triples conditioned on the previous generation.

As the rephrasing in large argument spans may introduce false-positive word noises, we employ a simple but effective masking strategy to ignore the impact of negative words while retaining the contribution of valuable correct words in the span.
For a triple $(a_1,p_r,a_2)$, we calculate the importance of each word in an argument $a_i$ based on its semantic matching score obtained from the arguments restoration algorithm. For those words which are recovered from the syntactic tree, we set them to the average value of other words. We finally normalize the reciprocals of these importance scores and randomly select 15\% of all words according to the probabilities distribution.
These sampled words will be masked to not calculate their gradients in training. Note that we only mask the words in arguments as the predicate is short and less noisy.

\section{Experiment}

This work proposes a syntactically robust training framework including two knowledge restoration strategies.
Therefore, our experiments are intended to demonstrate the effectiveness as well as the robustness of the proposed framework on test sets.

\subsection{Datasets}

We use the standard training set OpenIE4~\cite{kolluru-etal-2020-imojie}, and the constructed sets $\mathcal{D}^{\Phi}, \mathcal{D}^{\Psi}$ for model training. During evaluation, in addition to the benchmark dataset CaRB~\cite{bhardwaj2019carb}, we build a syntactically diverse evaluation set to validate the robustness of OpenIE model.

\subsubsection{Training set}

\begin{table}[hpt]
\begin{tabular}{|l|c|p{1.7cm}|p{1.7cm}|}
\hline
  \begin{minipage}{24pt}\ \\\textbf{Data}\end{minipage} & \begin{minipage}{1.7cm}\ \\\textbf{\# samples}\end{minipage} & \textbf{Fact-level accuracy} & \textbf{Span-level accuracy} \\
\hline
  $\mathcal{D}$ & 215,356 & \qquad / & \qquad / \\
\hline
  $\mathcal{D}^{\Phi}$ & 429,171 & \quad\ 87\% & \quad\ 34\% \\
\hline
  $\mathcal{D}^{\Psi}$ & 382,752 & \quad\ 91\% & \quad\ 71\% \\
\hline
\end{tabular}
\caption{Train set statistics and the human verification results. We randomly sample 100 samples for each dataset and evaluate two fine-grained metrics.}
\label{tab:train-set-statistics}
\end{table}

We use the dataset OpenIE4 as the basic set $\mathcal{D}$ in our experiment, which is published by~\cite{kolluru-etal-2020-imojie} and prep-processed by~\cite{kolluru-etal-2020-openie6}. The data is automatically built by running OpenIE-4, ClausIE, and RnnOIE on the sentences obtained from Wikipedia.

To estimate the quality of the generated samples of $\mathcal{D}^{\Phi}$ and $\mathcal{D}^\Psi$, we conduct fine-grained human verification by randomly samplling 100 data samples from each set. For a fair comparison, taking the triples from the human-annotated dataset CaRB as the reference criteria, we evaluate the generated samples on fact-level and span-level, respectively. Specifically, a triple is fact-level correct if all elements in the triple conform with the definition of arguments or predicate. A triple is span-level correct only if all arguments and predicates contain the complete words span in the sample sentence. The overall statistics are shown in Table~\ref{tab:train-set-statistics}.
We can see that though the fact-level accuracy shows the useable for $\mathcal{D}^{\Phi}$, the spans of arguments and predicate are extremely inaccurate with the accuracy of 34\%. By further performing the algorithms of syntactic tree walking-based arguments restoration and predicate restoration, we improve both the fact-level and span-level accuracy to 91\% and 71\%, suggesting the satisfaction of the generated data.

\subsubsection{Evaluation set}

\begin{table}[hpt]
\begin{tabular}{|l|p{1.1cm}|p{1.1cm}|p{1.1cm}|}
\hline
  \textbf{Data} & \textbf{\# sent.} & \textbf{arg}$_{.len}$ & \textbf{pre}$_{.len}$ \\
\hline
  CaRB & 1282 & 14.9 & 2.7 \\
\hline
  CaRB-AutoPara & 2269 & 17.3 & 2.3 \\
\hline
\end{tabular}
\caption{Evaluation set statistics. The \# sent. refers to the total number of sentences, and arg$._{len}$/pre$._{len}$ are the average lengths of argument/predicate of all samples in corresponding data, respectively.}
\label{tab:eval-set-statistics}
\end{table}

We use the standard benchmark CaRB~\cite{bhardwaj2019carb} to evaluate the proposed framework, which is a high-quality crowdsourced dataset with 1282 sentences and each sentence has manually annotated about 4 n-tuples.

In order to evaluate the syntactic robustness of OpenIE models, we build a syntactically diverse dataset based on CaRB with the proposed framework. We generate 5 paraphrases for each sentence from CaRB, and get 2269 high-quality sentences after performing the knowledge restoration. We refer to this automatically generated dataset as CaRB-AutoPara. The statistics of both datasets are shown in Table~\ref{tab:eval-set-statistics}. We can see that the newly built dataset is twice as large in scale and the lengths of arguments and predicates conform with the CaRB.

\subsection{Evaluation Metrics}

We use the scoring system proposed by~\cite{bhardwaj2019carb} to evaluate the OpenIE models on two test sets. The system first creates an all-pair matching table, with each column as a prediction tuple and each row as a gold tuple. It then computes single-match precision and multi-match recall by considering the number of common tokens in (gold, perdition) pair for each element of the fact.

Based on the confidence with each output triple, we report three important metrics: (1) Optimal F1: the largest F1 value in the P-R curve, (2) AUC: the area under the P-R curve, and (3) Last F1: the F1 score computed at the point of zero confidence.

\subsection{Experimental Settings}

We follow the original work to train a BART-based paraphrase model~\cite{sun2021aesop} on  ParaNMT-small~\cite{chen2019multi}, and the syntactic mapping set is collected from~\cite{wieting-gimpel-2018-paranmt}.
For knowledge restoration, we use the pretrained BERT~\cite{devlin-etal-2019-bert} model to calculate the cosine similarity, and fine-tune the T5 model~\cite{raffel2020exploring} with a language model head on it for the predicate restoration. The threshold $\tau$ and maintaining number of spans $k$ are empirically set to 0.7 and 5, respectively.

We train two implementations of our proposed framework based on the baseline model IMOJIE~\cite{kolluru-etal-2020-imojie} to investigate the effectiveness and syntactically robustness. IMOJIE$^\Phi$ is trained on $\mathcal{D}^\Phi$ that adopts the semantic similarity matching as the knowledge restoration method only. IMOJIE$^\Psi$ is trained on $\mathcal{D}^\Psi$ that uses the entire knowledge restoration algorithms.
All models followed the original implementations by using BERT as encoder and LSTM with the CopyAttention mechanism~\cite{cui-etal-2018-neural} as the decoder.
The detained parameters setting are shown in Appendix~\ref{append:parameters_settings}.

\subsection{Results on Different Datasets}

\begin{table}[hpt]
\centering
\begin{tabular}{p{1.6cm}>{\centering\arraybackslash}m{1.5cm}l>{\centering\arraybackslash}m{1.5cm}l>{\centering\arraybackslash}m{1.5cm}l}
\hline
  \begin{minipage}{1.6cm}\ \\\textbf{Model}\end{minipage} & \multicolumn{3}{c}{CaRB} \\
\cline{2-4}
         & F1 & AUC & Opt.F1 \\
\hline
  IMoJIE & 53.3 & 33.3 & 53.5 \\
  IMoJIE$^\Phi$ & 53.6 & 32.4 & 54.0 \\
  IMoJIE$^\Psi$ & \textbf{54.7} & \textbf{34.0} & \textbf{55.0} \\
\hline
\end{tabular}
\caption{Experimental results on CaRB.}
\label{tab:result-CaRB}
\end{table}
\textit{How does the proposed framework perform on the syntactic identically distributed data?}

In comparison with the baseline model, we find that the proposed syntactically robust training framework generally enhances the OpenIE model to achieve better performance on identically distributed data. As shown in Table~\ref{tab:result-CaRB}, we compare three models on the evaluation set CaRB, a minor scale dataset including 1282 human-annotated sentences. We can see that with the simple contextual similarity-based knowledge restoration, our model IMOJIE$^\Phi$ achieves better performance than the basic model on F1 and optimal F1 metrics. By training model with the entire knowledge restoration algorithms, the model IMOJIE$^\Psi$ outperforms the basic model by 1.4 F1 pts, 0.7 pts of AUC, and 1.5 pts of optimal F1. The results suggest that the OpenIE model is syntactic sensitive and can benefit from more syntactically sufficient training.

We argue that the CaRB data is the \textbf{syntactic identically distributed evaluation set} with the training set OpenIE4, as they are both sampled from a specific domain of Wikipedia, making them hold similar writing styles. For example, one sentence describes the fact of ``sb. won sth.'', and there are two sentences \textit{Murray Rothbard died in 1995 in Manhattan of a heart attack.} and \textit{Burnham died of heart failure at the age of 86, on September 1, 1947.} in the train and evaluation set respectively, where both sentences can extract triples with the same syntactic structure.

\noindent\textit{How does the proposed framework perform on a non-identically distributed datasets?}

\begin{table}[hpt]
\centering
\begin{tabular}{p{1.6cm}>{\centering\arraybackslash}m{1.5cm}l>{\centering\arraybackslash}m{1.5cm}l>{\centering\arraybackslash}m{1.5cm}l}
\hline
  \begin{minipage}{1.6cm}\ \\\textbf{Model}\end{minipage} & \multicolumn{3}{c}{CaRB-AutoPara} \\
\cline{2-4}
         & F1 & AUC & Opt.F1 \\
\hline
  IMoJIE & 51.1 & 31.4 & 51.2 \\
  IMoJIE$^\Phi$ & 52.6 & 32.1 & 52.8 \\
  IMoJIE$^\Psi$ & \textbf{53.4 } & \textbf{33.9} & \textbf{53.4} \\
\hline
\end{tabular}
\caption{Experimental results on CaRB-AutoPara.}
\label{tab:result-CaRBAuto}
\end{table}

\begin{table*}[ht]
\centering
\begin{tabular}{@{\extracolsep{-2pt}}c|llllllllll@{}}
\toprule[1.5pt]
   \textbf{Data} & \multicolumn{2}{c}{CaRB-C1} & \multicolumn{2}{c}{CaRB-C2} & \multicolumn{2}{c}{CaRB-C3} & \multicolumn{2}{c}{CaRB-C4} & \multicolumn{2}{c}{CaRB-C5} \\
\toprule[1.2pt]
   \textbf{Distance} & \multicolumn{2}{c}{0.227} & \multicolumn{2}{c}{0.386} & \multicolumn{2}{c}{0.468} & \multicolumn{2}{c}{0.541} & \multicolumn{2}{c}{0.665}  \\
\toprule[1.2pt]
   \textbf{Performance}   & AUC & Opt.F1 & AUC & Opt.F1 & AUC & Opt.F1 & AUC & Opt.F1 & AUC & Opt.F1 \\
\hline
  IMoJIE & 34.2 & 55.3 & 31.5 & 51.4 & 25.7 & 50.0 & 30.7 & 47.9 & 24.7 & 47.0 \\
  IMoJIE$^\Psi$ & \textbf{34.4} & \textbf{55.7} & 27.9 & \textbf{51.7} & \textbf{34.4} & \textbf{54.6} & \textbf{31.0} & \textbf{51.1} & \textbf{31.2} & \textbf{50.6} \\
\hline
\end{tabular}
\caption{Experimental results on different subjects of syntactic categories.}
\label{tab:results-carb-clusters}
\end{table*}

To investigate the effectiveness as well as syntactic robustness on open world setting, we evaluate models on the syntactically diverse set CaRB-AutoPara. We find that the proposed training framework comprehensively improves the syntactic robustness of the existing model, making it exhibit consistent better performance on no-identically distributed data. As shown in Table~\ref{tab:result-CaRBAuto}, the best performing model significantly outperforms the baseline by 2.3 F1 pts, 2.5 pts of AUC, and 2.2 pts of optimal F1. In contrast, the basic model shows a large degradation on this dataset compared to the original CaRB. The results suggest that our proposed syntactically robust training is more compatible with the open-world scenarios, and it is necessary to train and evaluate models on a non-identically distributed dataset.

The proposed evaluation set CaRB-AutoPara is more challenging for OpenIE models that are trained on existing general datasets. The syntactic structures are varied with respect to the training set. By taking the same example mentioned above,  there are sentences with a different voice and tense in the proposed CaRB-AutoPara, such as a question sentence \textit{Isn't it possible that he died of a heart attack?}.

\subsection{Analysis}
\label{subsec:analysis}

We further explore the performance of the model on different subsets representing prototypical syntactic categories, and analyze the trend of the model effect as the syntactic differences between the training set and the subset changed.

\paragraph{How to effectively measure the syntactic difference between sentences?}
As the training data is massive, we need an efficient metric of the syntactic differences between sentences to divide the test set and calculate the syntactic distance between the training set and test set.

\begin{figure}[hpt]
\includegraphics[scale=0.35]{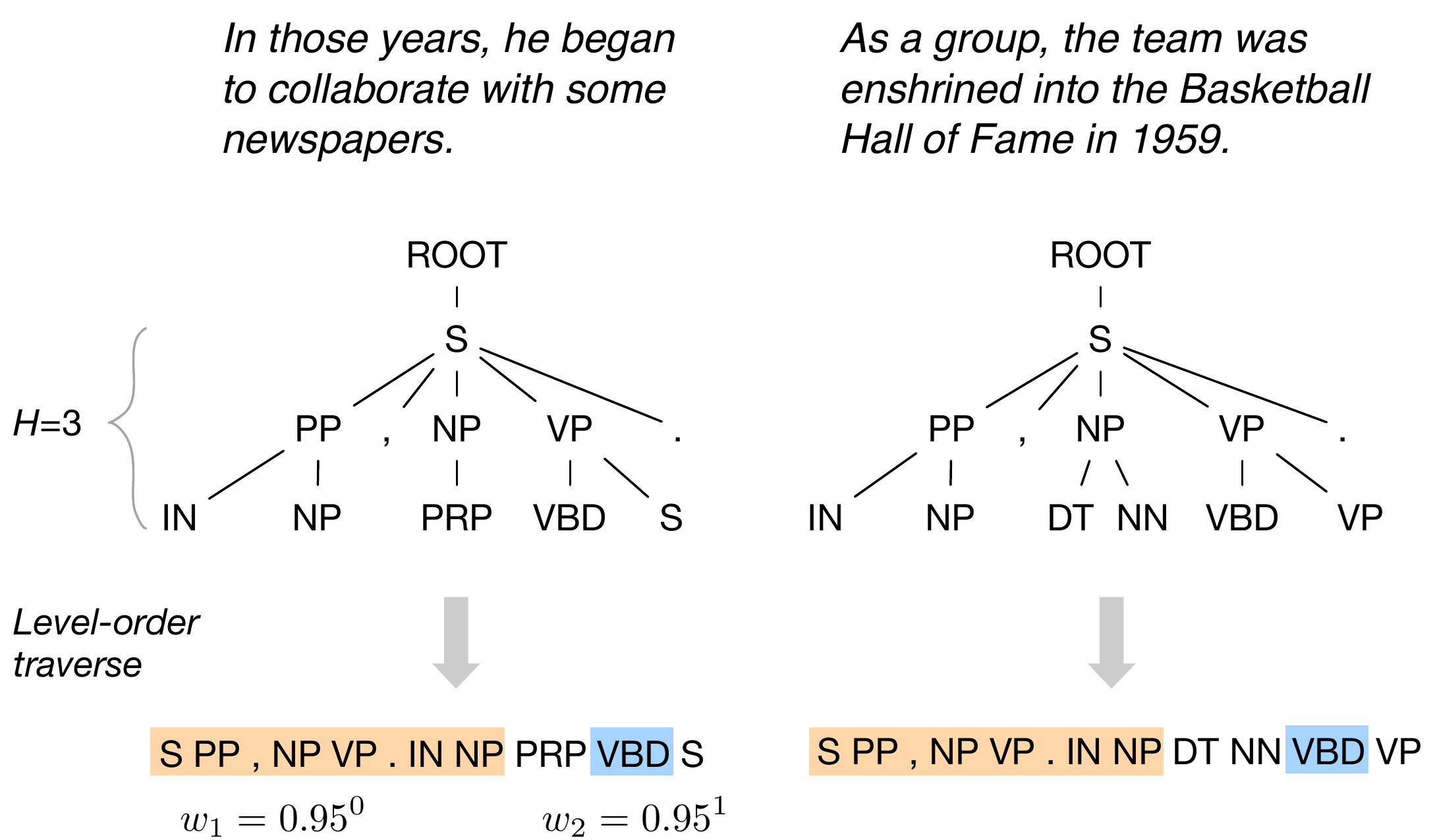}
\caption{Illustration of HW-Syntactic Distance.}
\label{fig:hw-syntactic-distance-example}
\end{figure}

We propose a simple but effective syntactic distance algorithm called Hierarchical Weighted Syntactic Distance (HW-Syntactic Distance), to measure the differences.
Intuitively, the more similar the skeleton of two sentences is, the less syntactic difference they have, i.e., the less syntactic distance.
We use a hierarchical weighted matching strategy on the constituency parse trees to calculate the syntactic distance between two sentences. As shown in Figure~\ref{fig:hw-syntactic-distance-example}, given two sentences with their constituency parse trees $T_1,T_2$ prune at height 3, we first transform the tree nodes in $T_1,T_2$ to sequences $q_1,q_2$ based on the level-order traversal. Then, we use the longest substring matching algorithm to accumulate the total matching length $l^{tot}$ of two sequences, where the length of $i$-th matched substring is multiplied by a sequentially discounting weight $w_i$. The final distance is a normalized value based on the minimum sequence length of $q_1,q_2$, and its value domain is [0, 1]. The detailed algorithm of HW-Syntactic distance is available in Appendix~\ref{append:subsec:hw-syntactic-distance}.

\begin{figure*}[h]
\centering
\includegraphics[scale=0.4]{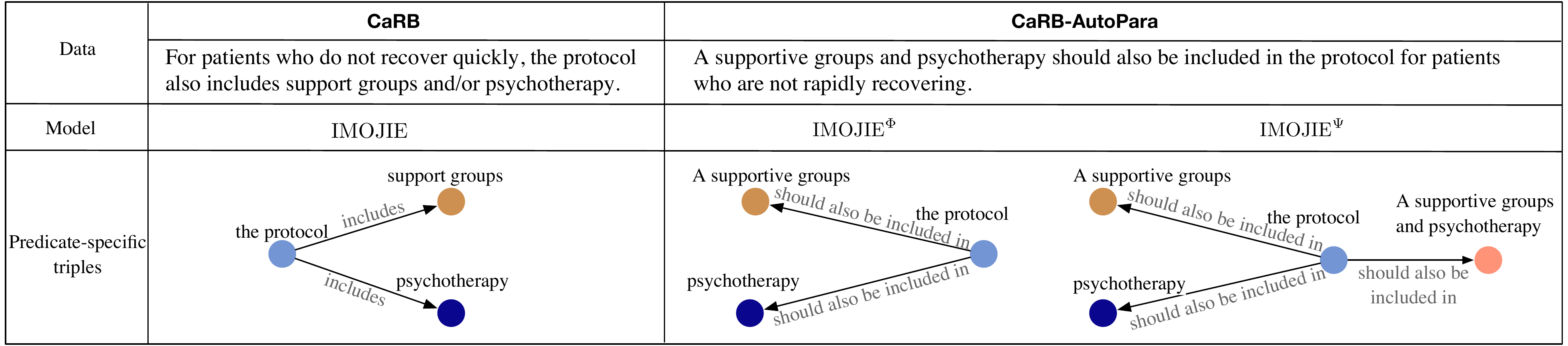}
\caption{A case study shows the partial predictions of model trained on the proposed framework.}
\label{fig:case-study}
\end{figure*}

\paragraph{How does the models trained on partial syntactic distribution perform on syntactic-specific data?}
\label{para:evaluation-subsets}

Based on this syntactic difference metric, we further analyze the performance of models trained on partially observed syntactic data $\mathcal{D}$ on different syntactic-specified datasets.

To this end, we first cluster the CaRB sentences into $k$ subsets with the metric of HW-Syntactic Distance\footnote{We use the K-means cluster algorithm, and cluster the samples with at most 300 epochs until convergence.}. Then, we randomly sample 300 sentences in the training set, and calculate the distance between the training set and each subset by averaging the distances among sampled training sentences and each cluster center.
We empirically clustered the CaRB sentences into 5 subsets with the optimal distance costs, and partial clustering results are available in Appendix~\ref{append:clustered-syntactic-samples}.

We find that the performance of the model on the subsets gradually increases as the syntactic distance between the training and test subsets decreases. As shown in Table~\ref{tab:results-carb-clusters}, compared to the best performance of 55.3 obtained on the subset CaRB-C5 with a distance of 0.227, the basic model only achieved an optimal F1 score of 47.0 on the subset CaRB-C1. In addition, we find that our fully enhanced model is consistently better than the basic model trained on partial syntactic distribution, suggesting that the proposed training framework improves the syntactic robustness of the OpenIE model comprehensively. We remain more analysis and results of syntactic distribution in Appendix~\ref{append:joint-words-distributions}.

\subsection{Case Study}

Figure~\ref{fig:case-study} shows the case study of our proposed framework with different implementations.
As is shown, compared to the original training sample, the generated sample exhibit a syntactically different structure. The model trained on the extended dataset with the semantic similarity-based knowledge restoration can only extract two separate triples around the predicate \textit{should also be included in}. By using the full knowledge restoration algorithms, the trained model can extract all related triples for the predicate. A part of generated samples based on the proposed syntactic robust training framework are shown in Appendix~\ref{append:syntactically-robust-samples}.

\section{Related Work}

\paragraph{Open Information Extraction} is a fundamental NLP task with a long research history~\cite{niklaus2018survey}. Traditional models adopt rule-based or statistical methods incorporating syntactic or semantic parsers to extract knowledge tuples~\cite{banko2007open,fader2011identifying,angeli-etal-2015-leveraging,del2013clausie,pal2016demonyms,saha2018open,stanovsky2015open,gashteovski2017minie}. Recently, neural models that either adopt sequence labeling strategies~\cite{stanovsky2018supervised,roy2019supervising,zhan2020span,kolluru-etal-2020-openie6,yu2021maximal}, or leverage sequence generative paradigms~\cite{cui2018neural,sun2018logician,kolluru-etal-2020-imojie} have achieved promising result.
To alleviate the problem that neural models rely heavily on labor-intensive annotated data, \cite{tang2020syntactic} proposes an unsupervised method that pretrains the model on synthetic data automatically labeled by patterns and then refines it using the RL process.

\paragraph{Paraphrase Generation} has proven to be useful for adversarial training and data augmentation~\cite{zhou2021paraphrase}. Early methods adopt hand-crafted rules~\cite{mckeown1983paraphrasing}, synonym substitution~\cite{bolshakov2004synonymous}, machine translation~\cite{quirk2004monolingual}, and deep learning~\cite{gupta2018deep,liu2019unsupervised} to improve the quality of generated sentences. To acquire syntactic diverse samples, recent studies involve reinforcement learning~\cite{qian2019exploring} or syntactic constrains~\cite{iyyer2018adversarial,goyal2020neural,sun2021aesop} into the models.

\section{Conclusion}

In this paper, we focus on solving the problem of partially observable of syntactic distribution on training data, and propose a syntactically robust training framework that enables OpenIE models to be trained on a syntactic-abundant distribution based on diverse paraphrase generation. We propose a knowledge restoration algorithm to recover the deformed triples in syntactically transformed sentences based on semantic similarity-based matching and syntactic tree walking. To investigate the syntactic robustness of models, we build a syntactically diverse evaluation set that is consistent with the real-world setting. The experimental result with extensive analysis demonstrated the efficiency of our framework.

\clearpage

\section*{Acknowledgement}
We thank all reviewers for their work and suggestions.
We thank Xiaozhi Wang for his help with insightful comments during this work.
This work is supported by the Key-Area Research and Development Program of Guangdong Province (2019B010153002), the NSFC Youth Project (62006136) and a grant from the Institute for Guo Qiang, Tsinghua University (2019GQB0003).

\section*{Limitations}
Although we have extensively studied different paraphrase generation models with diverse syntactic, it is difficult to guarantee the quality of the generated sentences in a specific domain. In this paper, some poorly generated sentences can cause errors to propagate into knowledge restoration and further lead to omitted triples. We built a syntactically diverse dataset to evaluate the robustness of the OpenIE models.  However, researchers willing to use this dataset need to be aware of the inevitable noises due to the automatic generation process.

\bibliography{custom}
\bibliographystyle{acl_natbib}

\clearpage
\appendix

\section{Model Parameters settings}
\label{append:parameters_settings}

We train all models on an NVIDIA Tesla V100 with 32GB GPU ARM. Hyperparameter settings for the paraphrase generation, knowledge restoration and OpenIE are listed in Table \ref{tab:setting_paraphrase_model}, \ref{tab:setting_knowledge_restore_model} and \ref{tab:setting_openie_model}, respectively.

\begin{table}[h]
    \centering
    \begin{tabular}{p{4.2cm}r}
      \toprule[1.5pt]
        \textbf{Hyperparameter} & \textbf{Value} \\
      \hline
        Backbone Model &  BART$_{base}$ \\
        Model Dimension &  768\\
        Learning Rate & 3e-5 \\
        Target Tree Height  & 2 \\
        Optimizer & Adam \\
      \hline
    \end{tabular}
    \caption{Settings for paraphrase generation model.}
    \label{tab:setting_paraphrase_model}
\end{table}

\begin{table}[h]
    \centering
    \begin{tabular}{lr}
      \toprule[1.5pt]
        \textbf{Hyperparameter} & \textbf{Value} \\
      \hline
        Contextual Similarity Model &  BERT$_{base}$ \\
        Threshold $\tau$ & 0.7 \\
        Maintaining Spans $k$ & 5\\
      \hline
        Predicate Restoration Model & T5$_{base}$ \\
        Model Dimension &  768\\
        Learning Rate & 1e-3 \\
        Optimizer & Adafactor \\
      \hline
    \end{tabular}
    \caption{Settings for knowledge restoration model.}
    \label{tab:setting_knowledge_restore_model}
\end{table}

\begin{table}[h]
    \centering
    \begin{tabular}{lr}
      \toprule[1.5pt]
        \textbf{Hyperparameter} & \textbf{Value} \\
      \hline
        Backbone Model & BERT$_{small}$ \\
        Model Dimension &  768\\
        Learning Rate & 2e-5 \\
        LSTM Hidden Dimension & 256 \\
        LSTM Word Embedding & 100 \\
        Optimizer & Adam \\
      \hline
    \end{tabular}
    \caption{Settings for OpenIE model.}
    \label{tab:setting_openie_model}
\end{table}

\section{Syntactic Distribution Analysis}
\label{sec:dataset_analysis}

\subsection{Hierarchical Weighted Syntactic Distance}
\label{append:subsec:hw-syntactic-distance}

The proposed Hierarchical Weighted Syntactic Distance (HW-Syntactic Distance) is shown in algorithm~\ref{alg:hw-syntactic-distance}. Given two sentences with their constituency parse trees $T_1,T_2$, the algorithm outputs their syntactic distance in [0, 1], where a smaller value means a closer distance. We first get their level-order traversal sequences $q_1, q_2$. Then we calculate their discounting weighted optimal matching length based on dynamic programming effectively. The final distance is a normalized value based on the minimum sequence length of $q_1,q_2$.

\begin{algorithm}[ht]
\renewcommand{\algorithmicrequire}{\textbf{Input:}}
\renewcommand{\algorithmicensure}{\textbf{Output:}}
\caption{\textbf{HW-Syntactic Distance}}\label{alg:hw-syntactic-distance}
\begin{algorithmic}[1]
\Require Constituency parses $T_1,T_2$ of sentences $s_1,s_2$, pruning height $h$, discount factor $\alpha$
\Ensure Syntactic distance $d$ between $s_1,s_2$
\State Get trees $T_1^h,T_2^h$ pruned at height $h$, and their level-order traversal sequences $q_1, q_2$
\State Initialize total length and count $l=0;m=0$
\State $\textrm{A[i][0]=1}$ if $q_1[i] == q_2[0], i=1,...,q_1._{len}$
\State $\textrm{A[0][j]=1}$ if $q_1[0] == q_2[j], j=1,...,q_2._{len}$
\For {$i=2 \rightarrow q_1.{len}$}
    \For {$j=2 \rightarrow q_2.{len}$}
        \If {$q_1[i] == q_2[j]$}
          \State  $\textrm{A}[i][j]=\textrm{A}[i-1][j-1]+1$
        \Else
          \State $\textrm{A}[i][j]=0$
          \If {$\textrm{A}[i-1][j-1] > 1$}
            \State  $l=\textrm{A}[i-1][j-1]\times\alpha^{m}$
            \State $m++$
          \EndIf
        \EndIf
    \EndFor
\EndFor
\If {$\textrm{A}[i-1][j-1] > 1$}
    \State $l=\textrm{A}[i-1][j-1]\times\alpha^{m}$
\EndIf
\State Return $1 - l/min(q_1._{len}, q_2._{len})$
\end{algorithmic}
\label{alg:hw-syntactic-distance}
\end{algorithm}

\subsection{Joint Words Distributions}
\label{append:joint-words-distributions}

\begin{figure}[ht]
    \centering
    \includegraphics[scale=0.38]{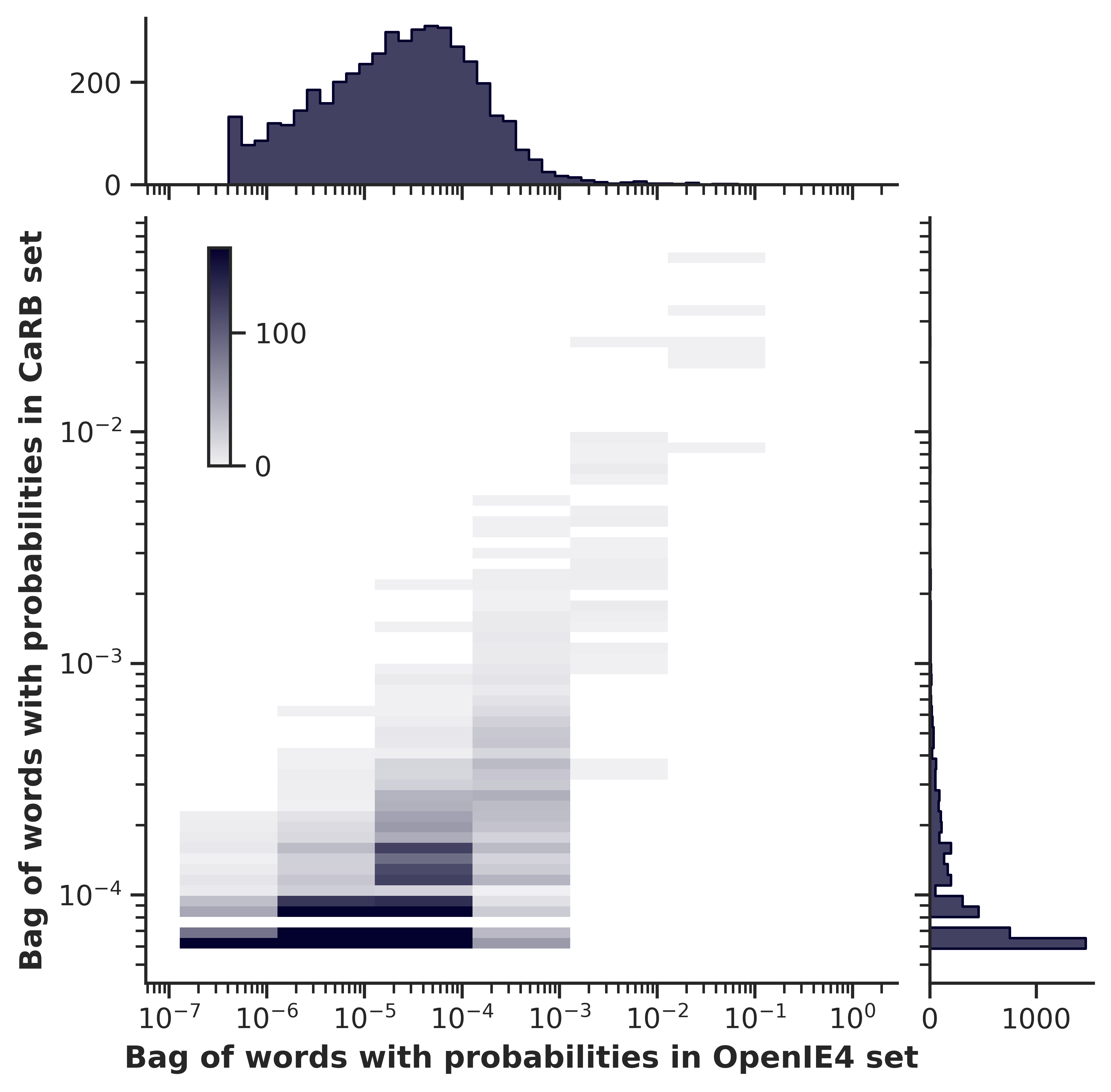}
\end{figure}

We analyze the joint probability distribution of distinct words between training data and CaRB data based on the vocabulary built on CaRB. As shown above, we find that there is a word-level distribution difference between the two datasets.

\subsection{Clustered Syntactic Samples}
\label{append:clustered-syntactic-samples}

We cluster the CaRB data with the HW-Syntactic Distance. Partial examples are shown in Table~\ref{append:tab-cluster-samples}.

\begin{table*}[hpt]
\centering
\renewcommand{\arraystretch}{2.2}
\begin{small}
\begin{tabular}{|c|l|c|}
\hline
  \textbf{Cluster} & \textbf{Syntactic Parse} & \textbf{Score} \\
\hline
  \multirow{3}{20pt}{\\ \ \\ \ \\ \ \\C1} & \begin{minipage}{12.5cm}
  In those years, he began to collaborate with some newspapers. \\ -- (ROOT (S (PP (IN ) (NP )) (, ) (NP (PRP )) (VP (VBD ) (S )) (. )))
  \end{minipage} & 0.245 \\
\cline{2-3}
  & \begin{minipage}{12.5cm}
  In Canada, there are two organizations that regulate university and collegiate athletics.  \\ -- (ROOT (S (PP (IN ) (NP )) (, ) (NP (EX )) (VP (VBP ) (NP )) (. )))
  \end{minipage} & 0.333 \\
\cline{2-3}
  & \begin{minipage}{12.5cm}
  As a result, it becomes clear that the microbe can not survive outside a narrow pH range.  \\ -- (ROOT (S (PP (IN ) (NP )) (, ) (NP (PRP )) (VP (VBZ ) (ADJP ) (SBAR )) (. )))
  \end{minipage} & 0.346 \\
\cline{2-3}
  & \begin{minipage}{12.5cm}
  However, during his rehearsal, Knievel lost control of the motorcycle and crashed into a cameraman.  \\ -- (ROOT (S (ADVP (RB )) (, ) (PP (IN ) (NP )) (, ) (NP (NNP )) (VP (VP ) (CC ) (VP )) (. )))
  \end{minipage} & 0.474 \\
\cline{2-3}
  & \begin{minipage}{12.5cm}
  If given this data, the Germans would be able to adjust their aim and correct any shortfall. \\ -- (ROOT (FRAG (SBAR (IN ) (S )) (. )))
  \end{minipage} & 0.711 \\
\hline
  \multirow{3}{20pt}{\\ \ \\ \ \\ \ \\C2} & \begin{minipage}{12.5cm}
  HTB 's aim is for an Alpha course to be accessible to anyone who would like to attend the course , and in this way HTB seeks to spread the teachings of Christianity .  \\ -- (ROOT (S (S (NP ) (VP )) (, ) (CC ) (S (PP ) (VP )) (. )))
 \end{minipage} &  0.019 \\
\cline{2-3}
  & \begin{minipage}{12.5cm}
  `` Business across the country is spending more time addressing this issue , '' says Sen. Edward Kennedy ( D. , Mass . ) .  \\ -- (ROOT (SINV (`` ) (S (NP ) (VP )) (, ) ("" ) (VP (VBZ )) (NP (NP ) (PRN )) (. )))
 \end{minipage} &  0.069 \\
\cline{2-3}
  & \begin{minipage}{12.5cm}
  Returning home , Ballard delivers her report , which her superiors refuse to believe .  \\ -- (ROOT (S (S (VP )) (, ) (NP (NNP )) (VP (VBZ ) (NP )) (. )))
 \end{minipage} &  0.131 \\
\cline{2-3}
  & \begin{minipage}{12.5cm}
  `` It 's really bizarre , '' says Albert Lerman , creative director at the Wells Rich Greene ad agency . \\ -- (ROOT (SINV (`` ) (S (NP ) (VP )) (, ) ("" ) (VP (VBZ )) (NP (NP ) (, ) (NP )) (. )))
 \end{minipage} &  0.136 \\
\cline{2-3}
  & \begin{minipage}{12.5cm}
  Feeling the naggings of a culture imperative , I promptly signed up . \\ -- (ROOT (S (S (VP )) (, ) (NP (PRP )) (VP (ADVP ) (VBD ) (PRT )) (. )))
 \end{minipage} &  0.210 \\
\hline
  \multirow{3}{20pt}{\\ \ \\ \ \\ \ \\C3} & \begin{minipage}{12.5cm}
   Historically , Aiseau was a village dedicated to agriculture , logging , but also to the industry . \\ -- (ROOT (S (NP (NNP )) (, ) (NP (NNP )) (VP (VBD ) (NP )) (. )))
 \end{minipage} &  0.015 \\
\cline{2-3}
  & \begin{minipage}{12.5cm}
  They beat Milligan 1-0 , Grand View 3-0 , Webber International 1-0 and Azusa Pacific 0-0 to win the NAIA National Championships . \\ -- (ROOT (FRAG (S (NP ) (VP )) (, ) (NP (NP ) (, ) (NP ) (CC ) (NP )) (. )))
 \end{minipage} &  0.051 \\
\cline{2-3}
  & \begin{minipage}{12.5cm}
  For the record , Jeffrey Kaufman , an attorney for Fireman 's Fund , said he was `` rattled -- both literally and figuratively . '' \\ -- (ROOT (S (PP (IN ) (NP )) (, ) (NP (NP ) (, ) (NP ) (, )) (VP (VBD ) (SBAR )) (. ) ("" )))
 \end{minipage} &  0.154 \\
\cline{2-3}
  & \begin{minipage}{12.5cm}
  Crouched at shortstop , Bert Campaneris , once Oakland 's master thief , effortlessly scoops up a groundball and flips it to second . \\ -- (ROOT (S (S (VP )) (, ) (NP (NP ) (, ) (ADVP ) (NP ) (, )) (VP (ADVP ) (VP ) (CC ) (VP )) (. )))
 \end{minipage} &  0.162 \\
\cline{2-3}
  & \begin{minipage}{12.5cm}
  Now Mr. Broberg , a lawyer , claims he 'd play for free . \\ -- (ROOT (S (ADVP (RB )) (NP (NP ) (, ) (NP ) (, )) (VP (VBZ ) (SBAR )) (. )))
 \end{minipage} &  0.178 \\
\hline
  \multirow{3}{20pt}{\\ \ \\ \ \\ \ \\C4} & \begin{minipage}{12.5cm}
  In the U.S. , more than half the PC software sold is either for spreadsheets or for database analysis , according to Lotus . \\ -- (ROOT (S (PP (IN ) (NP )) (, ) (NP (NP ) (VP )) (VP (VBZ ) (PP ) (, ) (PP )) (. )))
 \end{minipage} &  0.101 \\
\cline{2-3}
  & \begin{minipage}{12.5cm}
  It is part of the Surrey Hills Area of Outstanding Beauty and situated on the Green Sand Way .  \\ -- (ROOT (S (NP (PRP )) (VP (VBZ ) (NP )) (. )))
 \end{minipage} &  0.018 \\
\cline{2-3}
  & \begin{minipage}{12.5cm}
  This is the U.N. group that managed to traduce its own charter of promoting education , science and culture . \\ -- (ROOT (S (NP (DT )) (VP (VBZ ) (NP )) (. )))
 \end{minipage} &  0.138 \\
\cline{2-3}
  & \begin{minipage}{12.5cm}
  Of the self - starting vacuum cleaner , he says : `` Could be Cuddles , { Mrs. Stinnett 's dog } . '' \\ -- (ROOT (S (PP (IN ) (NP )) (: ) (S (VP )) (, ) (NP (PRP )) (VP (VBZ ) (: ) (`` ) (S )) (. ) ("" )))
 \end{minipage} &  0.210 \\
\cline{2-3}
  & \begin{minipage}{12.5cm}
  According to the 2010 census , the population of the town is 2,310 .  \\ -- (ROOT (S (PP (VBG ) (PP )) (, ) (NP (NP ) (PP )) (VP (VBZ ) (NP )) (. )))
 \end{minipage} &  0.271 \\
\hline
  \multirow{3}{20pt}{\\ \ \\ \ \\ \ \\C5} & \begin{minipage}{12.5cm}
  Sen. Mitchell is confident he has sufficient votes to block such a measure with procedural actions . \\ -- (ROOT (S (NP (NNP ) (NNP )) (VP (VBZ ) (ADJP )) (. )))
 \end{minipage} &  0.003 \\
\cline{2-3}
  & \begin{minipage}{12.5cm}
  Dr. Pim played for Ireland against England in 1892 , 1893 , 1894 and 1896 . \\ -- (ROOT (S (NP (NNP ) (NNP )) (VP (VBD ) (PP ) (PP )) (. )))
 \end{minipage} &  0.051 \\
\cline{2-3}
  & \begin{minipage}{12.5cm}
  From 1909 to 1912 , the Miami Canal was dug , bypassing the rapids at the head of the North Fork .   \\ -- (ROOT (S (PP (PP ) (PP )) (, ) (NP (DT ) (NNP ) (NNP )) (VP (VBD ) (VP )) (. )))
 \end{minipage} &  0.112 \\
\cline{2-3}
  & \begin{minipage}{12.5cm}
  Mrs. Marcos has n't admitted that she filed any documents such as those sought by the government .   \\ -- (ROOT (S (NP (NNP ) (NNP )) (VP (VBZ ) (RB ) (VP )) (. )))
 \end{minipage} &  0.114 \\
\cline{2-3}
  & \begin{minipage}{12.5cm}
  Hapoel Lod played in the top division during the 1960s and 1980s , and won the State Cup in 1984 .  \\  -- (ROOT (S (NP (NNP ) (NNP )) (VP (VP ) (, ) (CC ) (VP )) (. )))
 \end{minipage} &  0.203 \\
\hline
\end{tabular}
\caption{A part of samples in different clusters of CaRB data based on the HW-Syantactic distance. The score refers to the distance between current sentence and corresponding clustering center.}
\label{append:tab-cluster-samples}
\end{small}
\end{table*}

\begin{table*}[hbt]
\begin{small}
\renewcommand{\arraystretch}{1.2}
\begin{tabular}{|l|l|}
 \hline
  \textbf{Original sample}  & \textbf{Generated sample} \\
 \hline
  \multirow{3}{7.5cm}{
  \begin{minipage}{7.5cm}
    This finding indicated that organic compounds could carry current. \\
    \textit{(This finding, indicated that, organic compounds could carry current)}
  \end{minipage}}
   &
  \begin{minipage}{7.5cm}
    According to these results, organic compounds can carry the current. \\
    \textit{(organic compounds, can carry, the current)}
  \end{minipage}\\
 \cline{2-2}
 & \begin{minipage}{7.5cm}
    This finding has shown that organic compounds are capable of transmitting impulses. \\
    \textit{(This finding, has shown, that organic compounds are capable of transmitting impulses)} \\
    \textit{(organic compounds, are capable, transmitting impulses)}
  \end{minipage}\\
 \cline{2-2}
  & \begin{minipage}{7.5cm}
    That this finding has shown that organic compounds can be operated. \\
    \textit{(this finding, has shown, that organic compounds can be operated)} \\
    \textit{(organic compounds, can be operated, )}
  \end{minipage}\\
 \hline
  \multirow{3}{7.5cm}{
    \begin{minipage}{7.5cm}
    Regulations meant that the original sixth lap would be deleted and the race would be restarted from the beginning of said lap. \\
    \textit{(Regulations meant that, would be deleted, the original sixth lap)} \\
    \textit{(Regulations meant that, would be deleted, the race)}
  \end{minipage}}
   &
  \begin{minipage}{7.5cm}
    According to the rules, the original sixth round will be removed and the race will be re started at the beginning of the round. \\
    \textit{(the race, will be re started, at the beginning of the round)} \\
    \textit{(the original sixth round, will be removed, )}
  \end{minipage}\\
  \cline{2-2}
  & \begin{minipage}{7.5cm}
    The rules have made it possible to cancel the original sixth round and restart the race at the start of the round. \\
    \textit{(The rules, have made it possible, to cancel the original sixth round)} \\
    \textit{(The rules, have made it possible, restart the race at the start of the round)}
  \end{minipage}\\
\cline{2-2}
  & \begin{minipage}{7.5cm}
    But the rules stipulated that the original sixth round would be removed and the race to be re-started at the beginning of the round. \\
    \textit{(The rules, stipulated, that the original sixth round would be removed)} \\
    \textit{(the race, to be re started, at the beginning of the round)} \\
    \textit{(the original sixth round, would be removed)}
  \end{minipage}\\
\hline
    \multirow{3}{7.5cm}{
    \begin{minipage}{7.5cm}
     Maduveya Vayasu song from nanjundi kalyana was a track played during marriages for many many years in Kannada. \\
    \textit{(Maduveya Vayasu, is, a song)} \\
    \textit{(Maduveya Vayasu song, is from, anjundi kalyana)} \\
    \textit{(Maduveya Vayasu song, was a track played during, marriages)}
  \end{minipage}}
   &
  \begin{minipage}{7.5cm}
     The song of maduveya vayasu from nanjundi kalyana has been played in the marriage of many years in kannada.\\
    \textit{(The song of maduveya vayasu from nanjundi kalyana, has been played, in the marriage of many years in kannada)} \\
    \textit{(The song of maduveya vayasu, is from, nanjundi kalyana)}
  \end{minipage}\\
  \cline{2-2}
  & \begin{minipage}{7.5cm}
     Maduveya vayasu, the song of nanjundi kalyana has been played in many marriages throughout the country. \\
    \textit{(the song of nanjundi kalyana, has been played, in many marriages throughout the country)}
  \end{minipage}\\
\cline{2-2}
  & \begin{minipage}{7.5cm}
    When they were married, they played the song of maduveya vayasu from nanjundi kalyana. \\
    \textit{(they, played, the song of maduveya vayasu from nanjundi kalyana) \\
         (they, were married)} \\
    \textit{(the song, is from, nanjundi kalyana)}
  \end{minipage}\\
 \hline
\end{tabular}
\end{small}
\caption{A part of generated syntactically robust data samples based the proposed framework.}
\label{append:tab-robust-samples}
\end{table*}

\section{Syntactically Robust Samples}
\label{append:syntactically-robust-samples}

Base on the proposed framework, a part of generated samples are shown in Table~\ref{append:tab-robust-samples}.

\end{document}